%% file: main.tex
\documentclass{article}
\usepackage{url}
\usepackage{graphicx} 
\usepackage{amsmath}
\usepackage{amsfonts}
\usepackage{tabularx}
\usepackage{booktabs}
\usepackage{amsmath}
\usepackage[version=3]{mhchem}
\usepackage{siunitx}
\usepackage{adjustbox}
\usepackage{subfig}

\usepackage{PRIMEarxiv}
\usepackage[citestyle=numeric-comp, sortcites=true,sorting=none]{biblatex}
\title{Multi-Task Modeling for Engineering Applications with Sparse Data
}

\author{
Yigitcan Comlek*\\
GE Aerospace Research \\
Niskayuna, NY.\\
\And
R. Murali Krishnan* \\
GE Aerospace Research \\
Niskayuna, NY.\\
\And
 Sandipp Krishnan Ravi*\\
 GE Aerospace Research \\
Niskayuna, NY.\\
\And
Amin Moghaddas\\
Edison Welding Institute \\
Columbus, OH.\\
\And
Rafael Giorjao\\
Edison Welding Institute \\
Columbus, OH.\\
\And
Michael Eff \\
Edison Welding Institute \\
Columbus, OH.\\
\And
Anirban Samaddar \\
Argonne National Lab\\
Chicago, IL.\\
\And
Nesar S. Ramachandra \\
Argonne National Lab\\
Chicago, IL.\\
\And
Sandeep Madireddy \\
Argonne National Lab\\
Chicago, IL.\\
\And
Liping Wang \\
GE Aerospace Research\\
Niskayuna, NY.\\
\thanks{Equal Contribution}}

\date{\today}

\begin{document}

\maketitle


\begin{abstract}
Modern engineering and scientific workflows often require simultaneous predictions across related tasks and fidelity levels, where high-fidelity data is scarce and expensive, while low-fidelity data is more abundant. This paper introduces an Multi-Task Gaussian Processes (MTGP) framework tailored for engineering systems characterized by multi-source, multi-fidelity data, addressing challenges of data sparsity and varying task correlations. The proposed framework leverages inter-task relationships across outputs and fidelity levels to improve predictive performance and reduce computational costs. The framework is validated across three representative scenarios: Forrester function benchmark, 3D ellipsoidal void modeling, and friction-stir welding. By quantifying and leveraging inter-task relationships, the proposed MTGP framework offers a robust and scalable solution for predictive modeling in domains with significant computational and experimental costs, supporting informed decision-making and efficient resource utilization.

\end{abstract}


\section{Introduction}

Modern engineering and scientific workflows frequently require simultaneous prediction across related tasks and fidelity levels \cite{forrester2007multi,peherstorfer2018survey,perdikaris2017nonlinear,chen2024latent,comlek2025heterogeneous,lam2015multifidelity}. In such contexts, some outputs are scarce and expensive to obtain, while others are cheaper and more abundant. Multi-task Gaussian processes (MTGPs), also known as multi-output Gaussian processes, offer a principled Bayesian framework to exploit inter-task correlations, enabling knowledge sharing that improves predictive accuracy and reduces the demand for large high-fidelity datasets \cite{liu2023learning,boyle2004dependent,yu2007gaussian}. Over decades of development, MTGPs have been applied across diverse domains, including time series forecasting, multitask optimization, and multifidelity classification, demonstrating their broad utility wherever data cost asymmetries and cross-task dependencies are present \cite{shi2023process, lei2023prediction, wang2024joint, mehta2021adaptive, yang2021hybrid, cao2024transfer, bonilla2007multi}. The central motivation for MTGPs is to leverage dependencies among related tasks to enhance predictive quality when high-fidelity information is limited \cite{le2015cokriging}. For example, predicting an airfoil’s lift coefficient from limited, expensive high-fidelity computational fluid dynamics (CFD) simulations can benefit from correlating with sufficient low-fidelity simulations \cite{perdikaris2017nonlinear}. Recent work in joint multi-objective and multifidelity optimization has also utilized MTGPs to balance exploration and exploitation across tasks, improving predictive performance and decision-making by explicitly modeling relationships among outputs and fidelities \cite{wang2024joint}.

Methodologically, MTGPs extend single-output GP regression to vector-valued functions via matrix-valued covariance structures that encode cross-output relationships. Foundational contributions include multi-task GP prediction and collaborative multi-output formulations, which formalize shared information across tasks and provide scalable modeling approaches. A key building block is coregionalization: the intrinsic coregionalization model (ICM) and the linear model of coregionalization (LMC) define positive semidefinite cross-task covariance constructions, widely used in cokriging and multi-output GP literature \cite{alvarez2011computationally}. MTGP approach, the ICM employs a coregionalization matrix that represents inter-task dependencies, often parameterized through parameterized principal component analysis (PPCA) or related low-rank decompositions to maintain numerical stability \cite{alvarez2011computationally}. Important nuances include the autokrigeability effect: under specific settings (e.g., noiseless outputs and isotopic data), predictions with ICM can reduce to independent per-output predictions, effectively cancelling inter-task transfer. Beyond coregionalization, advances in multi-output kernels expand MTGP expressivity. Convolved multiple-output GPs induce dependencies by convolving latent processes with output-specific smoothing kernels, enabling structured, interpretable cross-output correlations. Spectral mixture kernels and their multi-output extensions capture rich periodic and nonstationary behaviors across outputs, while sparse approximations and variational methods improve scalability for large datasets \cite{alvarez2012kernels,wilson2013gaussian}. The semiparametric latent factor model (SLFM) provides another lens on LMC, mixing latent Gaussian processes through parametric matrices to yield correlated outputs; connections to regularization theory further ground multitask kernels for vector-valued functions. As applications expand, heterogeneity in input domains becomes increasingly common, tasks may be defined over different feature spaces or dimensionalities. Addressing this, recent work proposes heterogeneous stochastic variational LMC frameworks that learn domain mappings alongside GP parameters, enabling effective input alignment and residual modeling to capture deviations from prior mappings. This direction complements established multi-fidelity strategies such as recursive co-kriging and transfer learning, forming a broader toolkit for knowledge transfer across tasks and fidelities.

\begin{figure}[!ht]
    \centering
    \includegraphics[width=0.7\textwidth]{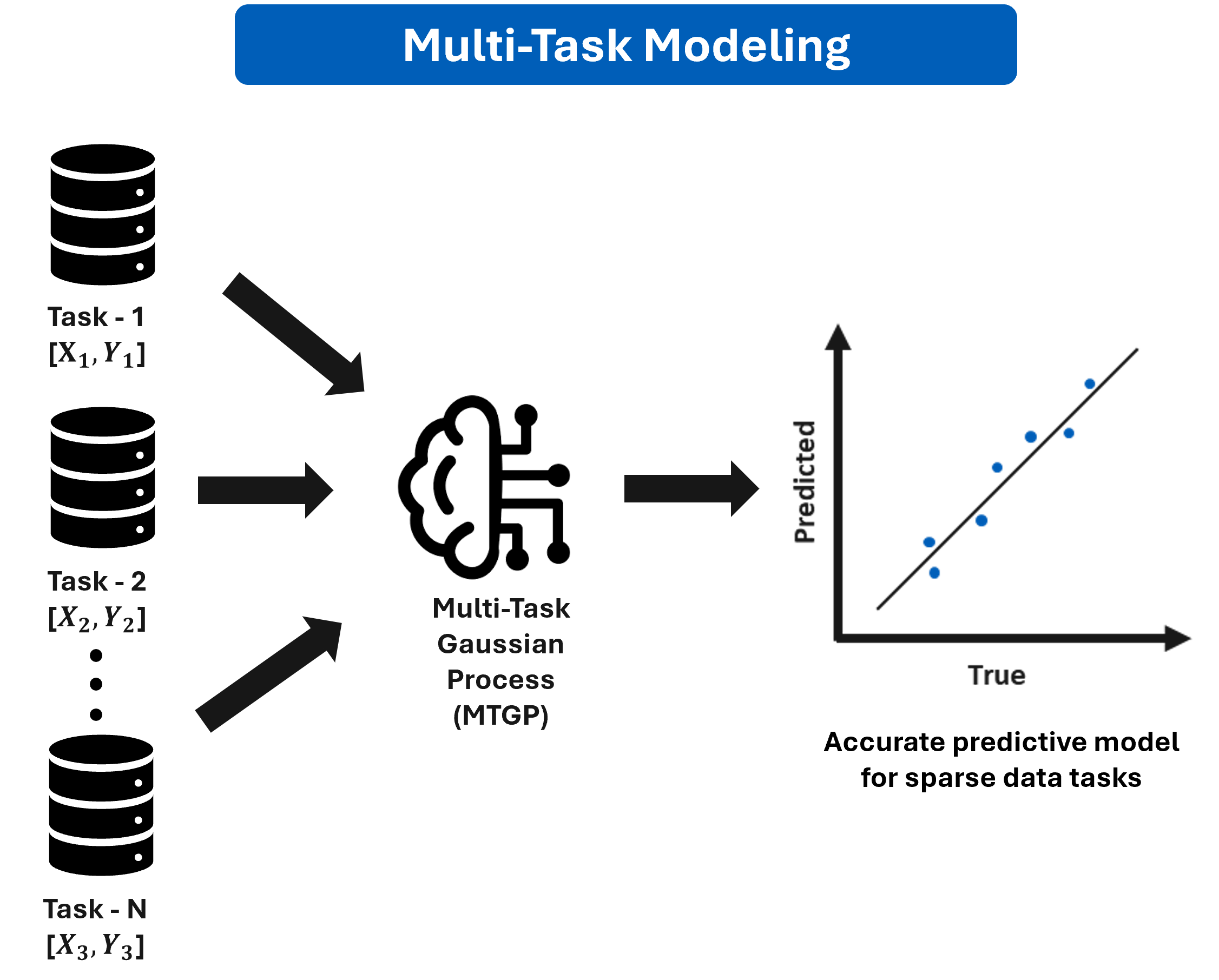}
    \caption{The multi-task modeling framework for sparse data applications}
    \label{fig:framework}
\end{figure}

This paper introduces a Multi-Task Gaussian Process (MTGP) framework tailored for engineering systems characterized by multi-source, multi-fidelity data, addressing challenges of data sparsity and varying task correlations. By leveraging inter-task relationships across outputs and fidelity levels, MTGP improves predictive accuracy and reduces reliance on costly high-fidelity data. The high-level schematic of the multi-task modeling framework is presented in Figure~\ref{fig:framework}. The framework is validated across three representative scenarios: the Forrester benchmark, which demonstrates MTGP’s ability to reduce error in primary-task predictions compared to single-task GP; 3D ellipsoidal void modeling, where MTGP achieves more accurate predictive outputs by integrating low-cost elastic simulations to predict expensive plastic stress outputs, reducing computational costs while maintaining accuracy; and friction-stir welding, where MTGP combines experimental distortion data with low-cost temperature simulations, achieving reduced errors and cost savings. Grounded in the theoretical foundations of coregionalization and enriched by advances in multi-output kernel design and sparse inference, our approach provides an actionable path for practitioners to deploy MTGPs in complex settings. By quantifying and leveraging inter-task relationships, the framework aims to deliver improved predictive performance and reduced data requirements, supporting informed decision-making in domains where computational and experimental costs are significant.

The paper is organized as follows: Section 1 introduces the motivation and background for multi-task Gaussian processes (MTGPs) in engineering systems, emphasizing their ability to leverage inter-task correlations to improve predictive accuracy and reduce reliance on high-fidelity data.  Section 2 outlines the methodology, including the formulation of MTGPs for vector-valued functions, coregionalization approaches, and advances in multi-output kernel design. Section 3 describes the analytical benchmark based on the Forrester function, extended to a two-task setting, and evaluates MTGP performance across varying correlation levels and data regimes. Section 4 presents the application of MTGPs to 3D ellipsoidal void modeling, highlighting the integration of elastic and plastic simulation tasks to improve predictions for computationally expensive outputs. Section 5 demonstrates the effectiveness of MTGPs in the friction-stir welding process, where experimental distortion data and low-cost temperature simulations are combined to achieve significant error reductions and cost savings. Finally, Section 6 concludes with a summary of contributions, emphasizing the scalability, robustness, and cost efficiency of MTGPs for predictive modeling in data-sparse, multi-fidelity engineering applications.

\input{method}
\input{benchmark}

\input{application_Void}
\input{application_PRISM}

\section{Conclusions} 

This work demonstrated that multi-task Gaussian processes (MTGP) provide a principled, data-efficient, and broadly applicable framework for engineering problems characterized by sparse observations, correlated outputs, and varying task fidelities. We began by formulating MTGPs for vector-valued functions with structured cross-task covariance, drawing on coregionalization approaches. These formulations separate input-space structure from output-task correlations, enabling the model to share information selectively across tasks while preserving task-specific behavior. Posterior inference in this framework generalizes scalar GP regression and yields joint predictions with calibrated uncertainty across all outputs, with hyperparameters learned via marginal likelihood maximization subject to appropriate positivity and positive-definiteness constraints. Collectively, these modeling choices allow MTGPs to exploit shared latent mechanisms that commonly arise in physical systems, improving data efficiency and robustness in multi-response settings.

We validated these capabilities on a controlled analytical benchmark based on the Forrester function, extended to a two-task setting with tunable inter-task correlation and variable data availability. This benchmark highlighted how MTGPs leverage auxiliary, correlated tasks to improve predictive accuracy on a primary task of interest. Across high, medium, and low correlation regimes, and under both low- and high-data conditions, MTGPs consistently matched or exceeded the performance of single-task GPs trained solely on the primary task. The gains were most pronounced when inter-task correlation was high and when additional data were available from the auxiliary task. To assess realism and engineering relevance, we applied MTGPs to a 3D ellipsoidal void modeling problem encompassing both elastic and plastic responses. This use case reflects a common industrial scenario in which multiple outputs arise from shared underlying mechanisms but are observed under different fidelities and computational costs. We considered two settings: a three-task elastic-only case to establish a same-fidelity baseline, and a mixed-fidelity case where low-cost elastic tasks assisted predictions for expensive plastic outputs. By jointly modeling outputs, the MTGP approach exploited cross-task and cross-fidelity structure to improve data efficiency and prediction quality, with RMSE reductions exceeding 50\%. Finally, we demonstrated the approach on a Friction-Stir Welding manufacturing process. Expensive high-fidelity experimental distortions were paired with lower-cost ABAQUS temperature simulations to form a multi-task learning problem. Under data-sparse experimental regimes, MTGP consistently improved  prediction accuracy versus a single-task GP trained only on experiments, achieving mean RMSE reductions of more than 15\%. These gains arise from learning cross-task structure between experiments and simulations, making MTGP an effective, cost-aware strategy for process modeling under realistic constraint.

In conclusion, this study demonstrates the significant advantages of MTGP for data-sparse scenarios, particularly when auxiliary information from related tasks is available. By incorporating relevant tasks into the modeling process, MTGP models achieve notable improvements in predictive performance, regardless of fidelity differences. This approach holds promise for a wide range of engineering and scientific problems, where data sparsity and multi-fidelity sources are common challenges. By enabling better predictions with limited data, MTGP offers a powerful tool for optimizing processes and improving decision-making in complex systems.

\section{Acknowledgment} 

GE Aerospace, Edison Welding Institute, and Argonne National Laboratory researchers are grateful for the support of the U.S. Department of Energy’s Office of Energy Efficiency and Renewable Energy (EERE) under the Advanced Manufacturing Office, Award Number DE-AC0206H11357. The views expressed herein do not necessarily represent the views of the U.S. Department of Energy or the United States Government.

\section{References}
\begin{center}\mbox{}\vspace{-\baselineskip}
    \printbibliography[heading=none]
\end{center}
\end{document}

%% file: method.tex
\section{Methodology}

\subsection{Gaussian processes for probabilistic regression}

Gaussian process (GP) regression provides a principled Bayesian framework for modeling unknown functions \cite{Rasmussen2006Gaussian}.
Unlike parametric models, which restrict the function space to a finite-dimensional family, GPs place a prior directly over functions, allowing them to adapt their complexity to the data.
Formally, a GP is a collection of random variables such that any finite subcollection has a join Gaussian distribution.
If
\begin{equation}
    f \sim \mathcal{GP}\big(m(\cdot),\, k(\cdot, \cdot)\big),
\end{equation}
then for any finite set of input points \(X=\{x_i\}_{i=1}^N\), the vector of function evaluations \((f(x_1),\ldots,f(x_N))^\top\) follows a joint Gaussian distribution with mean \(m(X)\) and covariance matrix \(K_{XX}\) defined by the kernel function \(k(\cdot,\cdot)\), i.e.,
\begin{equation}
    (f(x_1),\ldots,f(x_N))^\top \sim \mathcal{N}\big(m(X), K_{XX}\big), \qquad [K_{XX}]_{ij} = k(x_i, x_j)].
\end{equation}

The kernel plays a central role in determining properties of functions drawn from the prior, such as smoothness, periodicity, or stationarity.
Typical choices include the squared exponential, Matern, rational quadratic, and periodic kernels, as well as composite kernsl that can capture multi-scale and nonstationary function behaviors.

In regression problems, observations are assumed to be noisy:
\begin{equation}
    y_i = f(x_i) + \varepsilon_i,\qquad \varepsilon_i \sim \mathcal{N}(0,\sigma^2),
\end{equation}
where \(\sigma^2\) represents the noise variance.
Combining this likelihood with the GP prior yields a posterior GP
\begin{equation}
    f \mid X,Y \sim \mathcal{GP}\big(\bar m(\cdot), \bar k(\cdot, \cdot)\big),
\end{equation}
whose mean and covariance functions are
\begin{equation}
    \begin{aligned}
    \bar m(x) &= m(x) + k_{xX}\!\big(K_{XX} + \sigma^2 I_N\big)^{-1}\!\big(Y - m(X)\big), \\
    \bar k(x,x') &= k(x,x') - k_{xX}\!\big(K_{XX} + \sigma^2 I_N\big)^{-1}k_{Xx'}.
\end{aligned}
\end{equation}
Here, \(k_{xX}\) denotes the vector of kernel evaluations between the test input \(x\) and the training inputs \(X\), and \(I_N\) is the identity matrx of size \(N\).
These posterior expressions provide both the predictive mean (which serves as point estimates) and the predictive covariance (which quantifies uncertainty).
The ability to quantify uncertainty inherently in its calculus is a key strength of GP regression, especially in scientific and engineering applications where decision-making depends not only on predictions but also on their confidence levels~\cite{Rasmussen2006Gaussian}.

While classical GP regression is tailored to scalar-valued outputs, many problems involve multiple correlated outputs-for example, physical processes measured at multiple location, sensor fusion, or multi-physics systems where multiple fields interact.
Modeling each output independently ignores correlations between outputs, potentially leading to inefficient use of data and poorly calibrated uncertainty. MTGP address this shortcoming by explicitly modeling dependencies between multiple outputs.

\subsection{Multi-task learning with vector-valued Gaussian Process}
Consider a vector-valued function \(\mathbf{f}(x) = \big(f_1(x), f_2(x), \ldots, f_D(x)\big)^\top \in \mathbb{R}^D,\)
where \(D\) is the number of outputs or \emph{tasks}.
A multi-task GP places a joint Gaussian Process prior over these \(D\) functions:
\begin{equation}
    \mathbf{f} \sim \mathcal{GP}\big(\mathbf{m}(\cdot), \mathbf{K}(\cdot, \cdot)\big),
\end{equation}
where \(\mathbf{m}(x) \in \mathbb{R}^D\) is a vector of mean functions and \(\mathbf{K}(x,x') \in \mathbb{R}^{D \times D}\) is a matrix-valued kernel encoding both input-space correlations and cross-task dependencies.
For a set of \(N\) input points \(X = \{x_i\}_{i=1}^N\), the joint prior over all tasks and all inputs is
\begin{equation}
    \operatorname{vec}(\mathbf{F}) \sim \mathcal{N}\big(\operatorname{vec}(\mathbf{M}), \mathbf{K}(X,X)\big),
\end{equation}
where \(\mathbf{F}\in\mathbb{R}^{N\times D}\) contains the latent function evaluations,\(\mathbf{M}\) is the corresponding mean matrix, and \(\operatorname{vec}(\cdot)\) stacks columns into a vector.
Observations are modeled as
\begin{equation}
    \operatorname{vec}(\mathbf{Y}) = \operatorname{vec}(\mathbf{F}) + \varepsilon, \qquad \varepsilon \sim \mathcal{N}\big(0, \Sigma \otimes I_N\big),
\end{equation}
with \(\Sigma = \operatorname{diag}(\sigma_1^2, \ldots, \sigma_D^2)\) representing task-specific noise variances.
This formulation naturally generalizes standard GP regression and allows for posterior inference by replacing scalar kernels and covariance matrices with their block-structured multi-task counterparts.

Crucially, the structure of \(\mathbf{K}(x,x')\) determines how information is shared across outputs.
If the outputs are statistically independent, \(\mathbf{K}\) is block-diagonal.
However, in many physical systems, outputs are strongly coupled through shared underlying mechanisms where a structured \(\mathbf{K}\) enables the model to exploit these couplings, leading to improved data efficiency, and predictive performance.

\subsection{Coregionalization models multi-task modeling}
A widely used approach for constructing \(\mathbf{K}\) for multi-task modeling is through \emph{coregionalization}, which explicitly decomposes cross-task and input-space dependencies.
The \emph{Linear Model of Coregionalization} (LMC) expresses \(\mathbf{K}\), the matrix-valued kernel encoding as 
\begin{equation}
    \mathbf{K}(x, x') = \sum_{q=1}^Q \mathbf{B}^{(q)}\, k_q(x,x'), \qquad \mathbf{B}^{(q)} \succeq 0,
\end{equation}
where each scalar kernel \(k_q\) captures a distinct latent structure in the input space (e.g., smooth trends, periodic components, or short-scale variability), and each positive semidefinite matrix \(\mathbf{B}^{(q)} \in \mathbb{R}^{D \times D}\) encodes how strongly each latent structure is expressed in each output.
The resulting covariance matrix produced by the kernel encoding \(\mathbf{K}\) is guaranteed to be positive semidefinite, highly expressive, allowing for capturing rich patterns of inter-task dependence at the cost of complexity.
The general LMC formulation can combine low-rank and diagonal components for more intricate latent space correlations.
Therefore the coregionalization matrix can be written down as,
\begin{equation}
    \mathbf{B}^{(q)} = W^{(q)} {W^{(q)}}^\top + \mathrm{diag}(\gamma^{(q)}),
\end{equation}
where \(W^{(q)} \in \mathbb{R}^{D \times R_q}\) captures shared latent factors and \(\gamma^{(q)}\) encodes task-specific contributions.
This structure allows the model to capture both global trends shared across tasks and local variations specific to each output, making it well-suited to model complex physical processes~\cite{kanagawa2018gaussian}.

The \emph{Semiparametric Latent Factor Model} (SLFM) constrains each coregionalization matrix be of rank-one.
Because of which the coregionalization matrix, \(\mathbf{B}^{(q)}\) can be expressed as the outer product of two vectors.
In this case,
\begin{equation}
    \mathbf{B}^{(q)} = w_q w_q^\top, \qquad w_q\in\mathbb{R}^D
\end{equation}
leading to
\begin{equation}
    \mathbf{K}_{\mathrm{SLFM}}(x,x') = \sum_{q=1}^Q (w_q w_q^\top) k_q(x,x').
\end{equation}
This can be interpreted as a factor model in which each latent kernel defines a shared input structure and the factor loadings \(w_q\) determine how tasks depend on each factor.
SLFM reduces the number of parameters from \(\mathcal{O}(Q D^2)\) (for a full LMC) to \(\mathcal{O}(Q D)\), which is particularly advantageous when the number of taska \(D\) is large.
In this paper, we have studied the performance of SLFMs for multi task modeling as they strike the right balance between representational power against model complexity.

\subsection{Posterior inference and learning in Coregionalization models}
Having defined the LMC/SLFM kernels that couple tasks, we now state the exact inference and training recipe used throughout the experiments. 
For training inputs \(X\) and test inputs \(X_*\), with vectorized targets \(\mathbf{y}=\operatorname{vec}(\mathbf{Y})\), the joint prior is multivariate Gaussian with
\begin{equation}
    \mathbf{K}(X, X) = \sum_{q=1}^Q \mathbf{B}^{(q)} \otimes K_q(X,X),
\end{equation}
Conditioning on a noise covariance \(\Sigma \otimes I_N\) yields a vectorized standard GP posterior with  posterior mean and covariance expressed as:
\begin{equation}
\begin{aligned}
    \mathbb{E}[\mathbf{f}_* \mid \mathbf{y}] &= \mathbf{K}(X_*,X)\big(\mathbf{K}(X,X)+\Sigma\otimes I_N\big)^{-1}\big(\mathbf{y} - \mathbf{m}(X)\big), \\
    \operatorname{Cov}[\mathbf{f}_* \mid \mathbf{y}] &= \mathbf{K}(X_*,X_*) - \mathbf{K}(X_*,X)\big(\mathbf{K}(X,X) + \Sigma\otimes I_N\big)^{-1}\mathbf{K}(X, X_*).
\end{aligned}
\end{equation}
This formulation generalizes the Scalar valued GP Regression and enables joint prediction and uncertainty quantification across all tasks.
Hyperparameters of the kernel function \(\{k_q\}\), the elements of \(\{\mathbf{B}^{(q)}\}\) of factor loadings \(w_q\), and noise variances \(\{\sigma_d^2\}\) are typically estimated by maximizing the log-marginal likelihood \(\log p(\mathbf{y}\mid X, \theta) = -\tfrac{1}{2}\mathbf{y}^\top K_\theta^{-1}\mathbf{y} - \tfrac{1}{2}\log|K_\theta| - \tfrac{ND}{2} \log(2\pi),\)
where \(K_\theta = \mathbf{K}(X,X)  \Sigma\otimes I_N\).
In LMC models, Positive-definiteness of the coregionalization matrix (\(\mathbf{B}^{(q)} \succeq 0 \)) can be ensured via Chloesky parameterization or low-rank factorization, and positivity of kernel hyperparameters via log-transformations.
We use Adam-based gradient-based optimization scheme to enable efficient training even for moderately large \(N\) and \(D\).

The Friction-Stir Welding process studied in this work involves multiple physically coupled outputs observed at the same input configurations.
These outputs arise from common underlying mechanisms but display different levels of smoothness and characteristic scales.
Using a multi-task GP framework allows the model to share information across outputs when supported by the data while retaining flexibility to model differences across tasks.
Using a multi-task GP framework allows the model to share information across outputs supported by the data while retaining flexibility to model differences across tasks.
The LMC family of kernels, including ICM and SLFM, provides a principled and computationally tractable means of capturing these dependencies. By separating input and output correlations, MTGPs can achieve high data efficiency, improved uncertainty calibration, and robust generalization in multi-response problems~\cite{Alvarez2012}.

%% file: benchmark.tex
\section{Benchmark - Forrester Function}

The MTGP model is first applied to an analytical problem, the Forrester function~\cite{soton64699}, to demonstrate its functionality and applicability to multi-task systems. 
The Forrester function is a well-known mathematical benchmark often utilized in optimization and machine learning tasks due to its non-linear behavior and suitability for testing predictive models. 
It is a one dimensional function with a formulation defined as 
\begin{equation}
    f(x) = \mathbf{a}\,((6x - 2)^2 \, \sin(12x-4)) + \mathbf{b}\,(x-0.5)
    \label{eq:forrester}
\end{equation}
where $x$ is typically constrained between $[0,1]$ and the parameters $a$ and $b$ are coefficients that determine the scale and the form of the function. 
The function exhibits complex patterns, including multiple local minima and maxima, making it an ideal candidate for evaluating the performance of advanced modeling techniques, such as GPs or neural networks. 

In this study, the Forrester function is extended to a multi-task setting by introducing variations in output characteristics across tasks. 
The goal is to evaluate the predictive performance of the MTGP model on a specific task of interest compared to a single-task GP model built solely on that task. 
To achieve this, different instances of the Forrester function are created to represent distinct tasks, and these tasks are modeled together using MTGP. 
This approach allows the study to assess how effectively MTGP leverages information from related tasks to improve predictions for the primary task of interest.

\begin{figure}[!ht]
    \centering
    \includegraphics[width=0.7\textwidth]{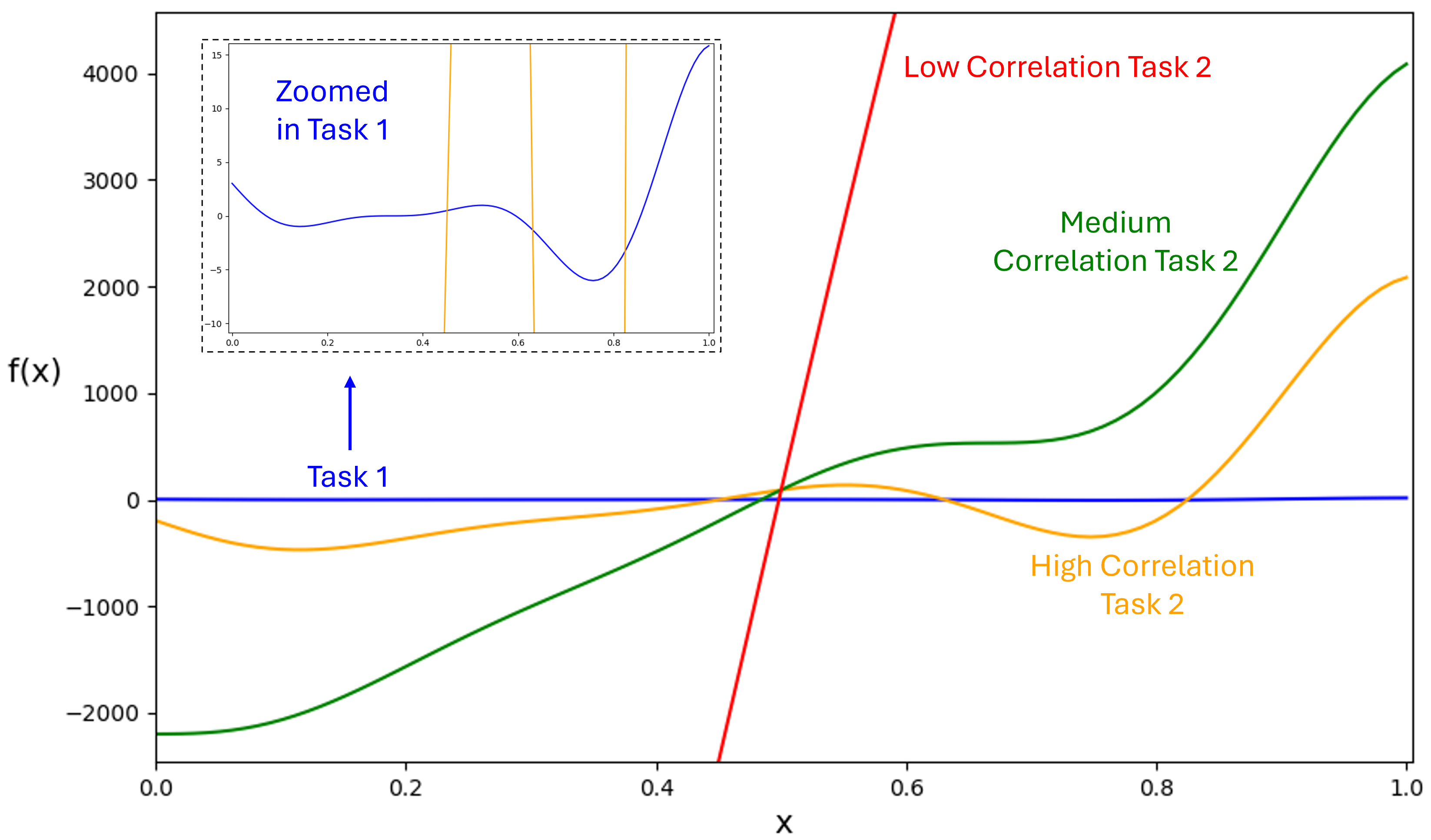}
    \caption{Low (r = 0.33), Medium (r = 0.55), and High (r = 0.89) correlated Forrester functions}
    \label{fig:forrester}
\end{figure}

To set up the problem, a two-task scenario is considered, where one task is designated as the primary focus for predictions (Task 1), while the other task (Task 2) serves as an auxiliary, correlated task. 
To account for diverse real-world scenarios, test cases are designed with varying levels of correlation between the tasks and different sample sizes per task. 
Specifically, three correlation levels based on Pearson correlation—high (r=0.89), medium (r=0.53), and low (r=0.33)—are examined to understand the impact of task relationships on model performance. 
Additionally, two data availability scenarios are considered: low data availability (5 samples per task) and high data availability (10 samples per task). The three correlation levels are also depicted in Figure \ref{fig:forrester}.
This setup provides a comprehensive framework for evaluating the robustness and adaptability of MTGP under different conditions.

For the modeling process, the single-task GP model is built exclusively on Task 1, while the MTGP model is trained on both Task 1 and Task 2 simultaneously. 
A separate testing set for Task 1 is created to evaluate the predictive accuracy of both models. 
By comparing the performance of the single-task GP and MTGP models, the study aims to quantify the benefits of multi-task learning in terms of error reduction and predictive reliability. 
The inclusion of Task 2 in the MTGP model allows it to exploit shared information and correlations between tasks, potentially leading to significant improvements in predictions for Task 1.

\begin{table}[htb]
  \centering
  \caption{MTGP and GP prediction performance comparison for medium correlation case (r=0.85)}
  \label{tbl:high_corr}
  \renewcommand{\arraystretch}{1.2}
  \begin{tabular}{@{}p{0.28\linewidth} p{0.32\linewidth} p{0.18\linewidth}@{}}
    \toprule
    \textbf{Task Pair} & \textbf{MTGP \textbackslash \space GP RMSE } & \textbf{\% Improvement} \\
    \midrule
    Low T1--Low T2   & 3.68 \textbackslash \space 4.44 & 17.1 \\
    Low T1--High T2  & 3.08 \textbackslash \space 4.44 & 77.48 \\
    High T1--Low T2  & 1.15 \textbackslash \space 1.47 & 21.77 \\
    High T1--High T2 & 0.84 \textbackslash \space 1.47 & 42.86 \\
    \bottomrule
  \end{tabular}
\end{table}

\begin{table}[htb]
  \centering
  \caption{MTGP and GP prediction performance comparison for medium correlation case (r=0.55)}
  \label{tbl:medium_corr}
  \renewcommand{\arraystretch}{1.2}
  \begin{tabular}{@{}p{0.28\linewidth} p{0.32\linewidth} p{0.18\linewidth}@{}}
    \toprule
    \textbf{Task Pair} & \textbf{MTGP \textbackslash \space GP RMSE } & \textbf{\% Improvement} \\
    \midrule
    Low T1--Low T2   & 4.08 \textbackslash \space 4.44 & 8.10 \\
    Low T1--High T2  & 4.00 \textbackslash \space 4.44 & 9.91 \\
    High T1--Low T2  & 1.18 \textbackslash \space 1.47 & 19.73 \\
    High T1--High T2 & 1.07 \textbackslash \space 1.47 & 27.21 \\
    \bottomrule
  \end{tabular}
\end{table}

\begin{table}[htb]
  \centering
  \caption{MTGP and GP prediction performance comparison for medium correlation case (r=0.31)}
  \label{tbl:low_corr}
  \renewcommand{\arraystretch}{1.2}
  \begin{tabular}{@{}p{0.28\linewidth} p{0.32\linewidth} p{0.18\linewidth}@{}}
    \toprule
    \textbf{Task Pair} & \textbf{MTGP \textbackslash \space GP RMSE } & \textbf{\% Improvement} \\
    \midrule
    Low T1--Low T2   & 4.27 \textbackslash \space 4.44 & 3.82 \\
    Low T1--High T2  & 4.33 \textbackslash \space 4.44 & 2.48 \\
    High T1--Low T2  & 1.19 \textbackslash \space 1.47 & 19.05 \\
    High T1--High T2 & 1.23 \textbackslash \space 1.47 & 16.33 \\
    \bottomrule
  \end{tabular}
\end{table}

The comparison results on the Task 1 testing set for three different correlation scenarios—high, medium, and low—are presented in Tables ~\ref{tbl:high_corr}, ~\ref{tbl:medium_corr}, ~\ref{tbl:low_corr}, respectively. The findings clearly demonstrate the impact of incorporating Task 2 data into the MTGP model, particularly in scenarios with high correlation between tasks. 
For the high-correlation case, the addition of Task 2 significantly enhances the model's predictive performance, as evidenced by the substantial percentage improvements. 
This highlights the ability of MTGP to effectively leverage strong task relationships to refine predictions for the primary task. 
Furthermore, the results reveal that the improvements become increasingly pronounced as additional Task 2 data is incorporated into the model, regardless of whether Task 1 data availability is low or high. 
In high Task 1 data scenarios, even though the single-task GP model already performs well, the inclusion of relevant information from Task 2 further boosts the predictive capability of the MTGP model, showcasing its ability to extract and utilize auxiliary task information to achieve superior results. 
For the medium-correlation case, notable improvements are observed across both low and high Task 1 data scenarios, demonstrating the versatility of MTGP in handling moderately related tasks. 
However, as the correlation between tasks decreases, the percentage improvement diminishes, reflecting the reduced influence of Task 2 on Task 1 predictions. 
Finally, in the low-correlation scenario, while the improvement percentages are less pronounced compared to the high- and medium-correlation cases, the addition of Task 2 still results in better predictive performance than the single-task GP model. 
This indicates that even when task relationships are weak, MTGP can still extract useful information from auxiliary tasks to enhance predictions. Importantly, across all correlation scenarios, the availability of more data for both Task 1 and Task 2 leads to significant improvements in model performance. 
This is attributed to the MTGP model's ability to learn and exploit the relationships between tasks more effectively as the data volume increases.

In conclusion, the application  to the Forrester function serves to demonstrate and exlplain the capabilities of MTGP in handling multi-task systems. 
Furthermore, this study highlights the advantages of MTGP in multi-task settings, particularly when tasks exhibit varying degrees of correlation and data availability. 
The results demonstrate that MTGP can effectively utilize auxiliary task information to enhance predictions for the primary task, outperforming single-task GP models in scenarios with high task correlation and limited data. 
This approach underscores the potential of MTGP as a powerful tool for multi-task modeling, offering improved accuracy and efficiency in applications where related tasks can provide valuable contextual information, paving the way for its application to more complex real-world problems.

%% file: application_Void.tex
\section{Ellipsoidal Voids}

Ellipsoidal void modeling is crucial for understanding and predicting the behavior of materials under various structural and mechanical conditions \cite{eshelby1957determination,pardoen2000extended,scheyvaerts2011growth}. Voids, which are imperfections or cavities within a material, significantly influence its mechanical properties, such as strength, durability, and deformation characteristics. Among these, 3D ellipsoidal voids are particularly important due to their complex geometry and prevalence in real-world applications, such as aerospace components, automotive parts, and structural materials. Accurate modeling of these voids is essential for optimizing material design, improving performance, and ensuring safety in critical applications.

\begin{figure}[!ht]
    \centering
    \includegraphics[width=0.7\textwidth]{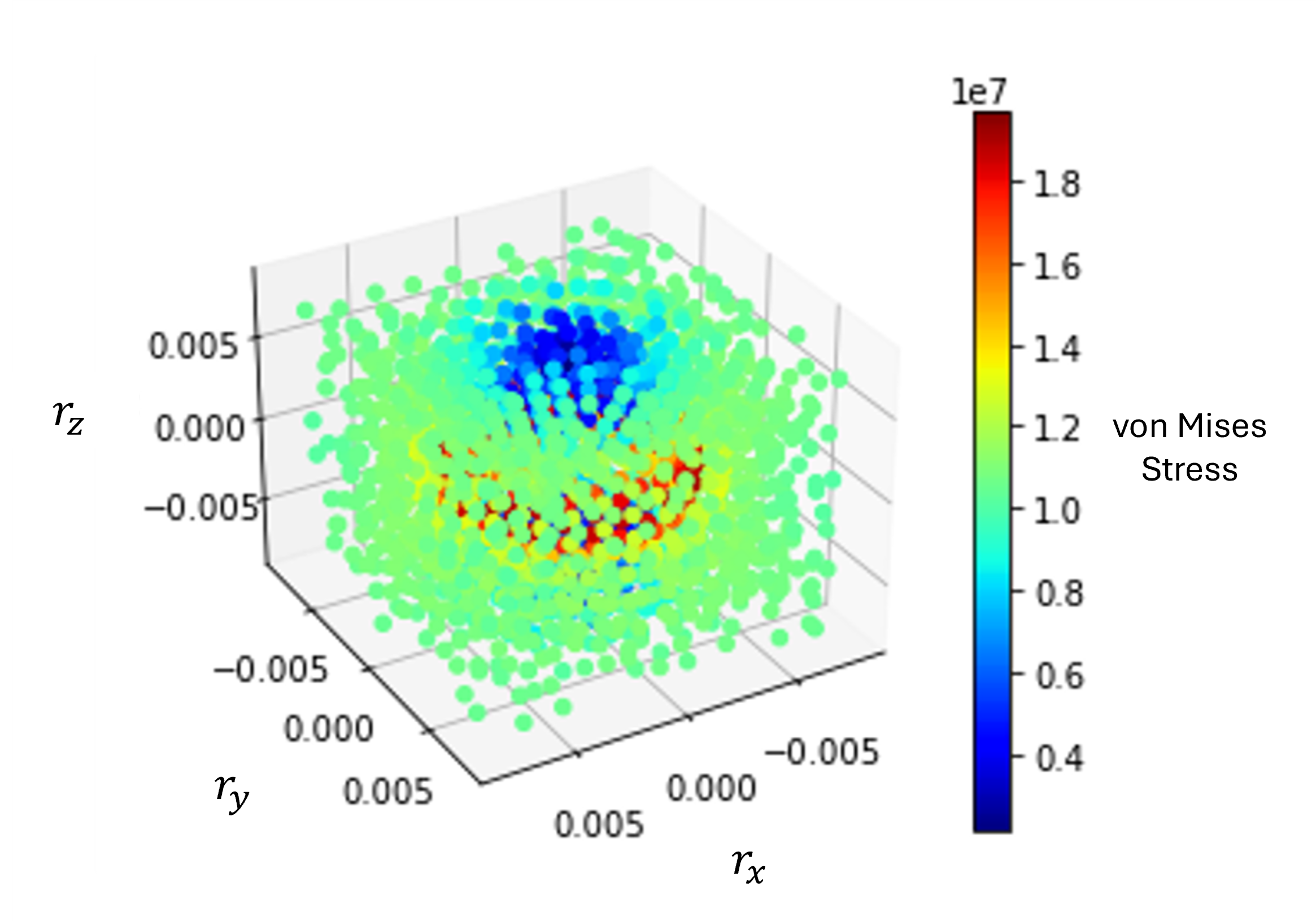}
    \caption{An example schematic of the ellipsoidal void system}
    \label{fig:void}
\end{figure}

This study focuses on the modeling of 3D ellipsoidal voids, where tasks differ in both outputs and fidelity of structural simulations, specifically elastic and plastic analyses. The schematic of the void problem is presented in Figure~\ref{fig:void}, providing a visual representation of the problem setup along with three design variables, $r_x$, $r_y$, and $r_z$, which define the radii of 3D ellipses within the material volume. Two distinct multi-task scenarios are analyzed to evaluate the effectiveness of MTGP in handling tasks with varying fidelities.

In the first scenario, a three-task modeling approach is employed, where all outputs are derived exclusively from elastic simulations. This serves as a baseline to evaluate predictive performance when tasks are homogeneous in fidelity. In the second scenario, a combination of tasks from both elastic and plastic simulations is utilized. The primary goal of this scenario is to improve predictions for the computationally expensive plastic simulation outputs by leveraging the low-cost and faster elastic simulation tasks. This approach demonstrates the potential of MTGP for multi-fidelity scenarios, enabling reduced computational overhead while maintaining high predictive accuracy. Table~\ref{tab:void_data} summarizes the specific tasks used in this study.

\begin{table}[htb]
  \centering
  \caption{Tasks used for modeling the Ellipsoidal Void case}
  \label{tab:void_data}
  \renewcommand{\arraystretch}{1.2}
  \begin{tabular}{@{}p{0.15\linewidth} p{0.40\linewidth}@{}}
    \toprule
    \textbf{Cases} & \textbf{Tasks}  \\
    \midrule
    Elastic   & Maximum Elastic Von Mises Stress (T1), Maximum Elastic Von Mises Strain (T2), Maximum Elastic Energy (T3) \\
    Plastic / Elastic   & Maximum Plastic  Von Mises Stress (T1), Maximum Elastic Von Mises Strain (T2),  Maximum Elastic Energy (T3) \\
    \bottomrule
  \end{tabular}
\end{table}


\subsection{Elastic Case}

For the all-elastic task case, a detailed study was conducted to compare the performance of MTGP against a single-task GP model. In this scenario, the maximum von Mises stress was designated as the primary task (Task 1 - T1), while the remaining two tasks—maximum von Mises strain (Task 2 - T2) and maximum elastic energy (Task 3 - T3)—were treated as auxiliary tasks providing additional information to support stress predictions. The study aimed to evaluate the impact of incorporating auxiliary task data on the predictive accuracy of the MTGP model.

\begin{figure}[!ht]
    \centering
    \includegraphics[width=1\textwidth]{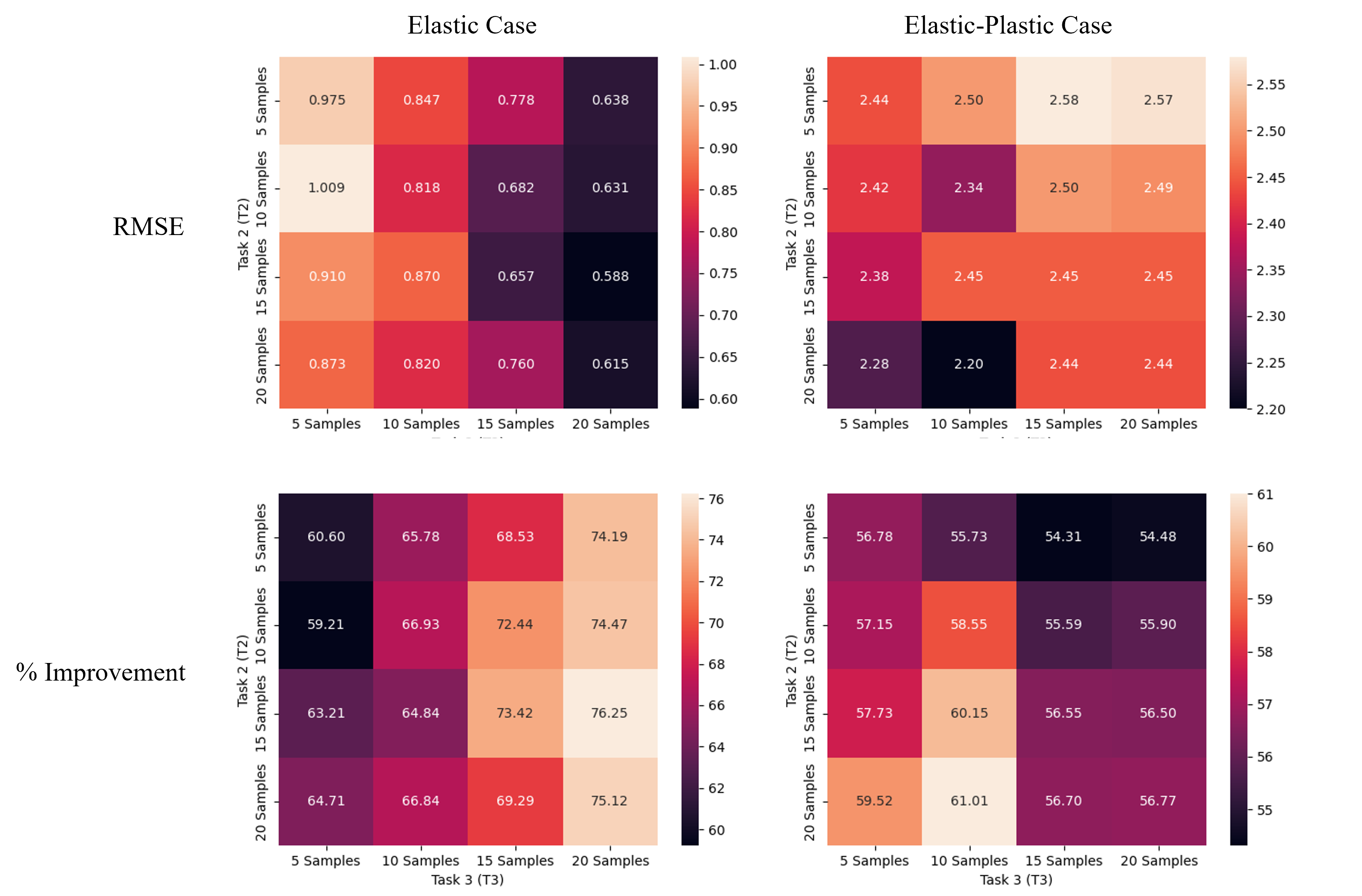}
    \caption{Predictive performance (RMSE and \% improvement) comparison between MTGP and GP for the elastic and elastic-plastic case studies on von Mises Stress (T1)}
    \label{fig:void_heatmaps}
\end{figure}

Figure~\ref{fig:void_heatmaps} illustrates the significant advantage of using MTGP over single-task GP for predicting maximum von Mises stress in the elastic case study. By leveraging shared information from auxiliary tasks, MTGP achieved substantial improvements in predictive performance, with RMSE values showing enhancements ranging from 59.21\% to 76.25\%. This improvement is attributed to MTGP's ability to exploit correlations between the primary task (maximum von Mises stress) and auxiliary tasks (maximum von Mises strain and maximum elastic energy), effectively integrating information across related outputs. The results from the elastic case study demonstrate consistent improvement in predictive accuracy with increased auxiliary data availability. The RMSE values for MTGP consistently decreased as the number of samples from auxiliary tasks (T2: maximum von Mises strain and T3: maximum elastic energy) increased, highlighting the importance of data availability in improving predictions for the primary task. For a fixed number of T2 samples, increasing the number of T3 samples led to significant improvements in RMSE. For example, with 5 T2 samples, increasing T3 samples from 5 to 20 reduced RMSE from $2.44$ MPa to $2.57$ MPa, corresponding to an improvement from $60.60\%$ to $74.19\%$. Similarly, for a fixed number of T3 samples, increasing the number of T2 samples also improved RMSE. For instance, with 5 T3 samples, increasing T2 samples from 5 to 20 reduced RMSE from $2.44$ MPa to $2.20$ MPa, corresponding to an improvement from $60.60\%$ to $64.71\%$. The highest improvement of $76.25\%$ was achieved when both T2 and T3 had 20 samples, with the RMSE reduced to $2.20$ MPa. This result highlights the synergistic effect of increasing data availability for both auxiliary tasks. The combination of increased samples from both T2 and T3 yielded the most substantial improvements, as demonstrated when T2 and T3 both had 20 samples, reducing the RMSE to $2.20$ MPa and achieving a $75.12\%$ improvement over single-task GP.

The MTGP model also demonstrated scalability, becoming increasingly effective as more data from auxiliary tasks was incorporated. This scalability allowed MTGP to refine predictions and improve accuracy as the volume of auxiliary task data increased. Even in data-sparse scenarios, MTGP consistently outperformed single-task GP, achieving a $60.60\%$ improvement with only 5 samples for both T2 and T3, showcasing its robustness in handling limited data availability. Furthermore, MTGP efficiently leveraged cross-task relationships by integrating low-cost elastic simulation data (T2 and T3), reducing reliance on high-fidelity data while maintaining high predictive performance. Elastic simulations, which are computationally inexpensive compared to plastic simulations, provided valuable contextual information such as strain and energy distributions that were closely correlated with plastic stress. By incorporating these auxiliary tasks, MTGP reduced the reliance on large datasets of high-fidelity plastic simulation outputs, which are often costly and time-consuming to generate.

In conclusion, the results from the elastic case study demonstrate the effectiveness of MTGP in leveraging auxiliary tasks to improve predictive accuracy for the primary task. The consistent reduction in RMSE values across varying sample sizes for T2 and T3 highlights the importance of data availability in multi-task learning. MTGP's ability to integrate low-cost elastic simulation data not only enhanced prediction quality but also reduced computational costs, making it a scalable and efficient solution for predictive modeling in engineering applications. These findings underscore the potential of MTGP to address multi-task and multi-fidelity challenges in data-sparse scenarios, enabling improved decision-making and optimization in complex systems.

\subsection{Elastic-Plastic Case}

For the elastic-plastic case study, a detailed investigation was conducted to evaluate the performance of MTGP against a single-task GP model. In this scenario, maximum plastic von Mises stress was designated as the primary task (Task 1 - T1), while maximum elastic von Mises strain (Task 2 - T2) and maximum elastic energy (Task 3 - T3) were treated as auxiliary tasks providing additional information to support stress predictions. The study aimed to assess the impact of integrating auxiliary task data on the predictive accuracy of the MTGP model.

Figure~\ref{fig:void_heatmaps} illustrates the significant advantage of using MTGP over single-task GP for predicting maximum plastic von Mises stress in the elastic-plastic case study. By leveraging shared information from auxiliary tasks, MTGP achieved substantial improvements in predictive performance, with RMSE values showing enhancements ranging from 54.31\% to 61.01\%. This improvement is attributed to MTGP's ability to exploit correlations between the primary task (maximum plastic von Mises stress) and auxiliary tasks (maximum elastic von Mises strain and maximum elastic energy), effectively integrating information across related outputs. The results from the elastic-plastic case study demonstrate consistent improvement in predictive accuracy with increased auxiliary data availability. The RMSE values for MTGP consistently decreased as the number of samples from auxiliary tasks (T2: maximum elastic von Mises strain and T3: maximum elastic energy) increased, highlighting the importance of data availability in improving predictions for the primary task. For a fixed number of T2 samples, increasing the number of T3 samples led to significant improvements in RMSE. For example, with 5 T2 samples, increasing T3 samples from 5 to 20 reduced RMSE from $2.44 \times 10$ MPa to $2.57 \times 10$ MPa, corresponding to an improvement from $56.78\%$ to $54.48\%$. Similarly, for a fixed number of T3 samples, increasing the number of T2 samples also improved RMSE. For instance, with 5 T3 samples, increasing T2 samples from 5 to 20 reduced RMSE from $2.44 \times 10$ MPa to $2.20 \times 10$ MPa, corresponding to an improvement from $56.78\%$ to $61.01\%$. 

The highest improvement of $61.01\%$ was achieved when T2 had 20 samples and T3 had 10 samples, with the RMSE reduced to $2.20 \times 10$ MPa. This result highlights the synergistic effect of increasing data availability for both auxiliary tasks. The combination of increased samples from both T2 and T3 yielded the most substantial improvements, as demonstrated when T2 and T3 both had 20 samples, reducing the RMSE to $2.44 \times 10$ MPa and achieving a $56.77\%$ improvement over single-task GP.

The MTGP model also demonstrated scalability, becoming increasingly effective as more data from auxiliary tasks was incorporated. This scalability allowed MTGP to refine predictions and improve accuracy as the volume of auxiliary task data increased. Even in data-sparse scenarios, MTGP consistently outperformed single-task GP, achieving a $56.78\%$ improvement with only 5 samples for both T2 and T3, showcasing its robustness in handling limited data availability. Furthermore, MTGP efficiently leveraged cross-task relationships by integrating low-cost elastic simulation data (T2 and T3), reducing reliance on high-fidelity data while maintaining high predictive performance. Elastic simulations, which are computationally inexpensive compared to plastic simulations, provided valuable contextual information such as strain and energy distributions that were closely correlated with plastic stress. By incorporating these auxiliary tasks, MTGP reduced the reliance on large datasets of high-fidelity plastic simulation outputs, which are often costly and time-consuming to generate.

In conclusion, the results from the elastic-plastic case study demonstrate the effectiveness of MTGP in leveraging auxiliary tasks to improve predictive accuracy for the primary task. The consistent reduction in RMSE values across varying sample sizes for T2 and T3 highlights the importance of data availability in multi-task learning. MTGP's ability to integrate low-cost elastic simulation data not only enhanced prediction quality but also reduced computational costs, making it a scalable and efficient solution for predictive modeling in engineering applications. These findings underscore the potential of MTGP to address multi-task and multi-fidelity challenges in data-sparse scenarios, enabling improved decision-making and optimization in complex systems.

%% file: application_PRISM.tex
\section{Modeling the Friction Stir Welding Process}

\begin{figure}[!ht]
    \centering
    \includegraphics[width=0.7\textwidth]{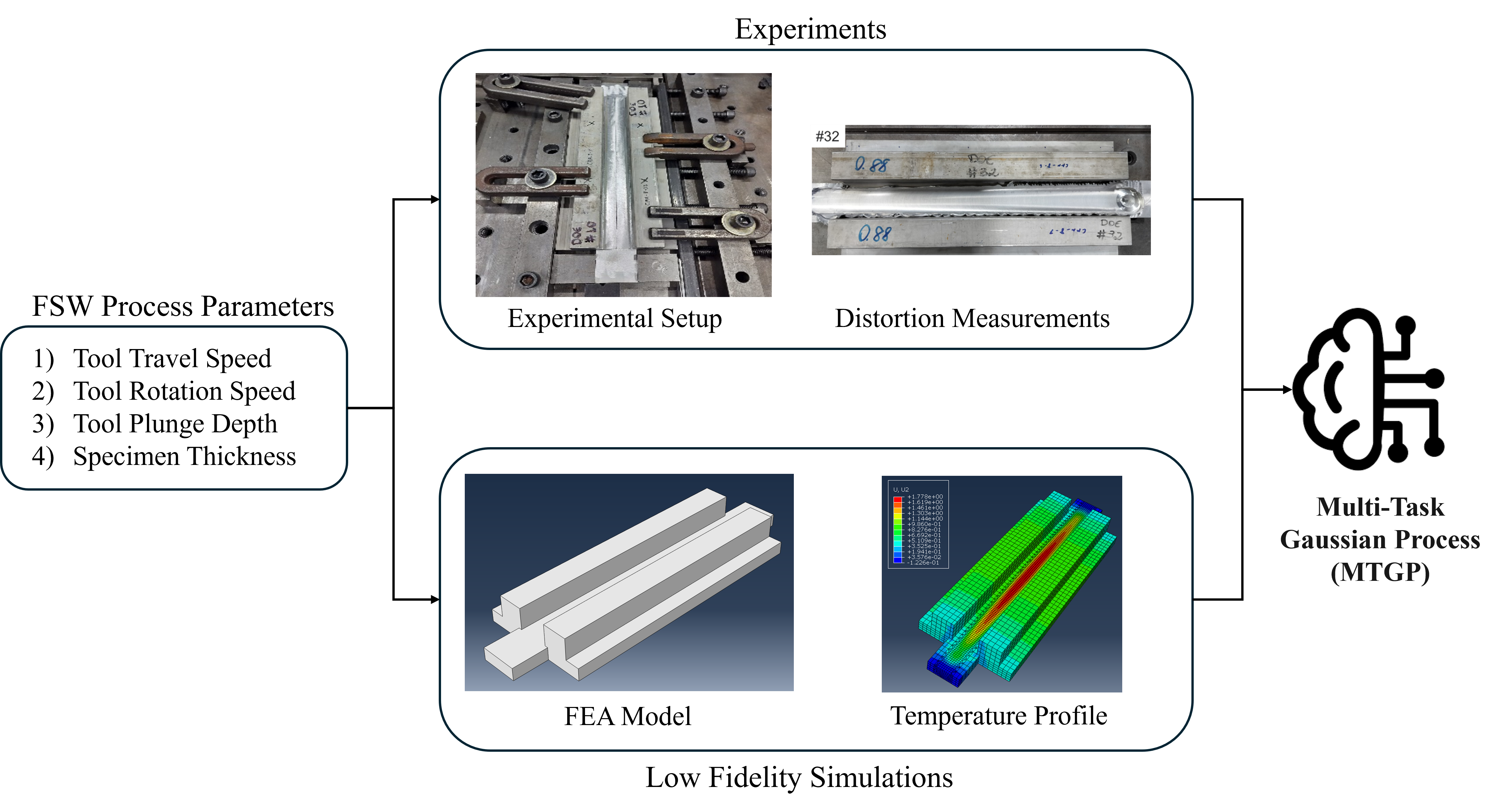}
    \caption{Friction stir welding modeling framework}
    \label{fig:fsw_schematic}
\end{figure}

Friction-Stir Welding (FSW) is an advanced solid-state joining process that has revolutionized the manufacturing of metallic components \cite{vilacca2011friction,thomas2003friction,sidhu2012friction, venu2019review}. Unlike traditional fusion welding techniques, FSW joins metals through the heat generated by friction between a rotating tool and the specimen. This localized heat softens the material, enabling the tool to stir and forge the metals together without melting them. FSW offers significant advantages, such as reduced defects, improved mechanical properties, and enhanced joint strength. Despite its advantages, FSW introduces distortions in the welded specimens due to the complex interplay of thermal, mechanical, and material behaviors during the process \cite{kulkarni2013residual}. These distortions can compromise the dimensional accuracy and structural integrity of the welded components, making it essential to understand and predict their occurrence. Figure~\ref{fig:fsw_schematic} provides a schematic representation of the FSW processes, highlighting the key parameters that influence distortions.

The distortions observed in post-weld samples are influenced by tool rotation speed, travel speed, plunge depth, and tool geometry. These parameters interact in a highly nonlinear manner, making it challenging to predict distortions accurately using traditional regression modeling approaches. Popular parametric regression methods, such as random forests and neural networks, are often unsuitable for this task due to their reliance on large datasets and inability to provide uncertainty estimates in predictions.

Data sparsity is a common issue in manufacturing and engineering applications, where the cost of conducting experiments is high (each experiment costing thousands of dollars). This high cost limits the number of experimental samples that can be collected, making data sparsity a significant challenge in modeling the FSW process. To address these challenges, this study aims to develop a robust predictive model that captures the intricate relationships between process parameters and the resulting distortions. By leveraging a multi-task learning approach, the model seeks to reduce the complexity of predicting distortions in the welded specimen for the given set of process parameters, utilizing additional lower-cost tasks.

\subsection{Data Sources}

The process experts at Edison Welding Institute (EWI) can control the FSW system through four primary process parameters: tool rotation speed, tool travel speed, plunge depth of the tool, and the thickness of the welded specimen. These parameters play a critical role in determining the quality and distortions of the welded material, making their precise control essential for optimizing the FSW process. To model the complex relationships between these parameters and the resulting distortions, we utilize two distinct tasks, as summarized in Table~\ref{tab:data_sources}.

\begin{table}[htb]
  \centering
  \caption{Tasks used to model the FSW process}
  \label{tab:data_sources}
  \renewcommand{\arraystretch}{1.2}
  \begin{tabular}{@{}p{0.26\linewidth} p{0.33\linewidth} p{0.18\linewidth} p{0.15\linewidth}@{}}
    \toprule
    \textbf{Task} & \textbf{Description} & \textbf{Cost} & \textbf{Output} \\
    \midrule
    T1: Experimental Trials & FSW experiment on a single specimen & \$1,000s & Distortion (in) \\
    T2: Low-Fidelity Simulations & ABAQUS simulation with 23k mesh elements & 6 hours of CPU time & Temperature (K) \\
    \bottomrule
  \end{tabular}
\end{table}

The first task involves experimental trials, which provide high-fidelity data on the distortions observed in the welded specimens. These experimental outputs represent the primary task of interest for modeling, as they offer the most accurate representation of the physical process. Experimental trials involve conducting FSW on actual specimens under controlled conditions, where the process parameters such as tool rotation speed, travel speed, plunge depth, and specimen thickness are varied systematically. The distortions in the welded specimens are measured using high-precision instruments, ensuring the reliability and accuracy of the data. However, experimental trials are expensive and time-consuming, with each trial costing thousands of dollars. This high cost limits the number of samples that can be collected, making data sparsity a significant challenge in modeling the FSW process. Despite these limitations, experimental trials remain the gold standard for understanding the physical phenomena underlying the FSW process.

The second task is derived from low-fidelity simulations, which use a coarse mesh with reduced integration elements (DC3D8R and C3D8R) to approximate the FSW process. These simulations are computationally cheaper and faster to execute, requiring approximately six hours of CPU time per simulation. While low-fidelity simulations are less accurate compared to experimental trials, they capture valuable trends and relationships within the FSW process, making them a useful complementary data source. Low-fidelity simulations are particularly effective in modeling the thermal and mechanical behavior of the FSW process. They provide insights into temperature distributions, stress fields, and deformation patterns, which are critical for understanding the factors that contribute to distortions. By incorporating these outputs into the predictive model as auxiliary tasks, the low-fidelity simulations can enhance the model's ability to generalize and predict distortions under varying process conditions.

The challenge lies in effectively combining these two tasks to build a predictive model that balances accuracy and computational efficiency. By leveraging the outputs from low-fidelity simulations as auxiliary tasks, we aim to enhance the predictions for the high-fidelity experimental trials. This multi-task learning approach allows us to exploit the shared structure and correlations between the tasks, improving the model's ability to generalize and predict distortions under varying process conditions.

\subsection{Modeling Results with MTGP}

For data-sparse scenarios, GP models are often the preferred choice for modeling systems due to their ability to provide accurate predictions with limited data. However, this study explores the advantages of MTGP over traditional GP models by demonstrating the effectiveness of incorporating relevant auxiliary tasks to improve predictive performance. In this comparison, the GP model is built solely on experimental data, while the MTGP model integrates temperature outputs from low-fidelity simulations as an additional task alongside distortion outputs from the experimental data source. As part of the ongoing collaboration with EWI, 32 experimental trials and 32 low-fidelity simulations were conducted to provide the necessary data for this study. The experimental trials involved FSW of metallic specimens under controlled conditions, where process parameters such as tool rotation speed, travel speed, plunge depth, and specimen thickness were systematically varied. These trials provided high-fidelity data on distortions observed in the welded specimens, representing the primary task of interest.

In addition to experimental trials, low-fidelity simulations were performed. These simulations captured temperature distributions and other thermal trends within the FSW process. While computationally cheaper and faster to execute (approximately six hours of CPU time per simulation), low-fidelity simulations provided valuable contextual information that complemented the experimental data. The combination of these two tasks enabled the development of a robust predictive model that balances accuracy and computational efficiency. The 32 samples for both tasks were obtained through full factorial sampling within the design space, ensuring comprehensive coverage of the parameter combinations. To evaluate the performance of the models under data-sparse conditions, different scenarios were considered, with the number of training samples from the experimental task ranging between 4 and 16, and testing samples ranging between 28 and 16, respectively. In contrast, all 32 low-fidelity simulations were utilized in the MTGP modeling process to maximize the auxiliary information available. The results of this comparison are summarized in Table~\ref{tab:prism_results}, illustrating the improved predictive performance of the MTGP model compared to the GP model.

\begin{table}[htb]
  \centering
  \caption{Predictive performance comparison between MTGP and GP models}
  \label{tab:prism_results}
  \renewcommand{\arraystretch}{1.2}
  \begin{tabular}{@{}p{0.45\linewidth} p{0.32\linewidth} p{0.18\linewidth}@{}}
    \toprule
    \textbf{\# of Experimental Training \textbackslash \space Testing Samples} & \textbf{Mean MTGP \textbackslash \space GP RMSE (mm)} & \textbf{\% Improvement} \\
    \midrule
    4 \textbackslash \space 28 Samples  & 0.0034 \textbackslash \space 0.0040 & 15.00 \\
    8 \textbackslash \space 24 Samples  & 0.0033 \textbackslash \space 0.0041 & 19.51 \\
    12 \textbackslash \space 20 Samples  & 0.0032 \textbackslash \space 0.0039 & 17.95 \\
    16 \textbackslash \space 16 Samples  & 0.0031 \textbackslash \space 0.0041 & 24.40 \\
    \bottomrule
  \end{tabular}
\end{table}

The study reveals that even in extremely low-data scenarios, such as when only 4 experimental samples are available, the MTGP model significantly outperforms the traditional GP model. This improvement is primarily attributed to the inclusion of temperature data from low-fidelity simulations, which provide valuable contextual information to support predictions. As the number of experimental samples increases, the benefits of the MTGP model become even more pronounced. With more experimental data, the MTGP model is better able to learn the correlations and relationships between tasks, resulting in substantial improvements in predictive accuracy. This demonstrates the scalability of the MTGP approach, as it becomes increasingly effective with the availability of additional data.

The MTGP model demonstrates several key advantages in data-sparse scenarios. First, it consistently outperforms the traditional GP model, achieving reductions in RMSE ranging from 15\% to 24.40\%. This improvement is primarily attributed to the integration of auxiliary tasks, such as temperature outputs from low-fidelity simulations, which provide valuable contextual information to enhance predictions for the primary task. Second, the MTGP approach exhibits remarkable scalability, becoming increasingly effective as more experimental data becomes available. This scalability is particularly beneficial in engineering applications where data acquisition is both expensive and time-consuming. Third, the MTGP model offers significant cost efficiency by incorporating low-fidelity simulations, which require only six hours of CPU time compared to the thousands of dollars needed for experimental trials. This integration reduces the reliance on costly experimental data while maintaining high predictive accuracy. Finally, the MTGP model showcases robustness in handling fidelity differences, effectively combining information from high-fidelity experimental data and low-fidelity simulations. This versatility makes MTGP a powerful tool for addressing multi-task scenarios with varying levels of fidelity, enabling improved predictive performance and efficient modeling in complex engineering systems.

In conclusion, this study demonstrates the significant advantages of using MTGP for data-sparse scenarios, particularly when auxiliary information from related tasks is available. By incorporating relevant tasks into the modeling process, MTGP models achieve notable improvements in predictive performance, regardless of fidelity differences. This approach holds promise for a wide range of engineering and scientific problems, where data sparsity and multi-fidelity sources are common challenges. By enabling better predictions with limited data, MTGP offers a powerful tool for optimizing processes and improving decision-making in complex systems.